\documentclass[10pt,conference]{IEEEtran}
\IEEEoverridecommandlockouts
\pdfoutput=1
\usepackage{cite}
\usepackage{multirow}
\usepackage{booktabs}
\usepackage{algorithmic}
\usepackage{textcomp}
\usepackage{tabularx}
\usepackage{graphicx}  % for resizing
\usepackage{xcolor}
\usepackage{float}
\usepackage{hyperref}
\usepackage{tikz}

\def\BibTeX{{\rm B\kern-.05em{\sc i\kern-.025em b}\kern-.08em
    T\kern-.1667em\lower.7ex\hbox{E}\kern-.125em}}

\begin{document}

\title{Knowledge Graph Based Repository-Level Code Generation}
\author{
    \IEEEauthorblockN{Mihir Athale\IEEEauthorrefmark{1} and Vishal Vaddina\IEEEauthorrefmark{2}} \\
    \IEEEauthorblockA{\IEEEauthorrefmark{1}Computer Science, Northeastern University, Boston, MA, USA \\
    athale.m@northeastern.edu} \\
    \IEEEauthorblockA{\IEEEauthorrefmark{2}Applied Research, Quantiphi Inc., Toronto, ON, Canada \\
    vishal.vaddina@quantiphi.com} \\
    \thanks{\IEEEauthorrefmark{1}Work done during an internship at Quantiphi Inc.} \\
}

\maketitle

\begin{abstract}
Recent advancements in Large Language Models (LLMs) have transformed code generation from natural language queries. However, despite their extensive knowledge and ability to produce high-quality code, LLMs often struggle with contextual accuracy, particularly in evolving codebases. Current code search and retrieval methods frequently lack robustness in both the quality and contextual relevance of retrieved results, leading to suboptimal code generation. This paper introduces a novel knowledge graph-based approach to improve code search and retrieval leading to better quality of code generation in the context of repository-level tasks. The proposed approach represents code repositories as graphs, capturing structural and relational information for enhanced context-aware code generation. Our framework employs a hybrid approach for code retrieval to improve contextual relevance, track inter-file modular dependencies, generate more robust code and ensure consistency with the existing codebase. We benchmark the proposed approach on the Evolutionary Code Benchmark (EvoCodeBench) dataset, a repository-level code generation benchmark, and demonstrate that our method significantly outperforms the baseline approach. These findings suggest that knowledge graph based code generation could advance robust, context-sensitive coding assistance tools.
\end{abstract}


\section{Introduction}
The landscape of software development has fundamentally transformed with the emergence of large language models (LLMs) such as GPT-4\cite{openai2024gpt4technicalreport} and Claude-3.5-Sonnet, which demonstrate remarkable capabilities in translating natural language requirements into functional code. These models, trained on vast repositories of code and documentation, achieve unprecedented performance on standardized benchmarks like HumanEval\cite{chen2021evaluatinglargelanguagemodels}, showcasing their ability to generate syntactically correct and functionally appropriate code across diverse programming tasks. This advancement represents a significant milestone in automated programming assistance, promising to accelerate software development cycles and reduce the cognitive load on developers.

\par The integration of LLMs into software development workflows has seen widespread adoption through interfaces like ChatGPT, Claude AI, and various copilot tools\cite{10109345}. This adoption is driven by their ease of use, extensive knowledge base, and versatility in handling diverse coding tasks. The high-volume adoption demonstrates AI's positive impact on developer productivity, particularly in day-to-day development tasks.

\par However, several critical limitations emerge in professional software development environments. LLMs often generate code in isolation, lacking an understanding of project-specific architectural patterns\cite{dou2024whatswrongcodegenerated} and internal dependencies. This contextual disconnect leads to significant time investment in adapting generated code to project requirements. Additionally, LLMs frequently produce redundant code\cite{hou2024largelanguagemodelssoftware}, duplicating existing functionality and increasing maintenance complexity. Code style inconsistency presents another challenge, as LLMs generate code with varying styles\cite{wang2024functionalcorrectnessinvestigatingcoding}, often misaligned with project-specific conventions.

\par Current mitigation approaches primarily rely on enhancing code retrieval through web searches, documentation analysis, and pattern matching\cite{Di_Grazia_2023}\cite{tan2024promptbasedcodecompletionmultiretrieval}. While Retrieval Augmented Generation (RAG)\cite{NEURIPS2020_6b493230} systems show promise, traditional retrieval methods struggle with complex scenarios involving multiple interacting components. Recent agentic workflows\cite{li2024reviewprominentparadigmsllmbased}\cite{nguyen2024agilecoderdynamiccollaborativeagents} attempting iterative refinement face challenges in computational overhead and potential unproductive iteration loops.

\begin{figure}
    \centering
    \includegraphics[width=1\linewidth, trim=6cm 2.5cm 6cm 2.5cm, clip]{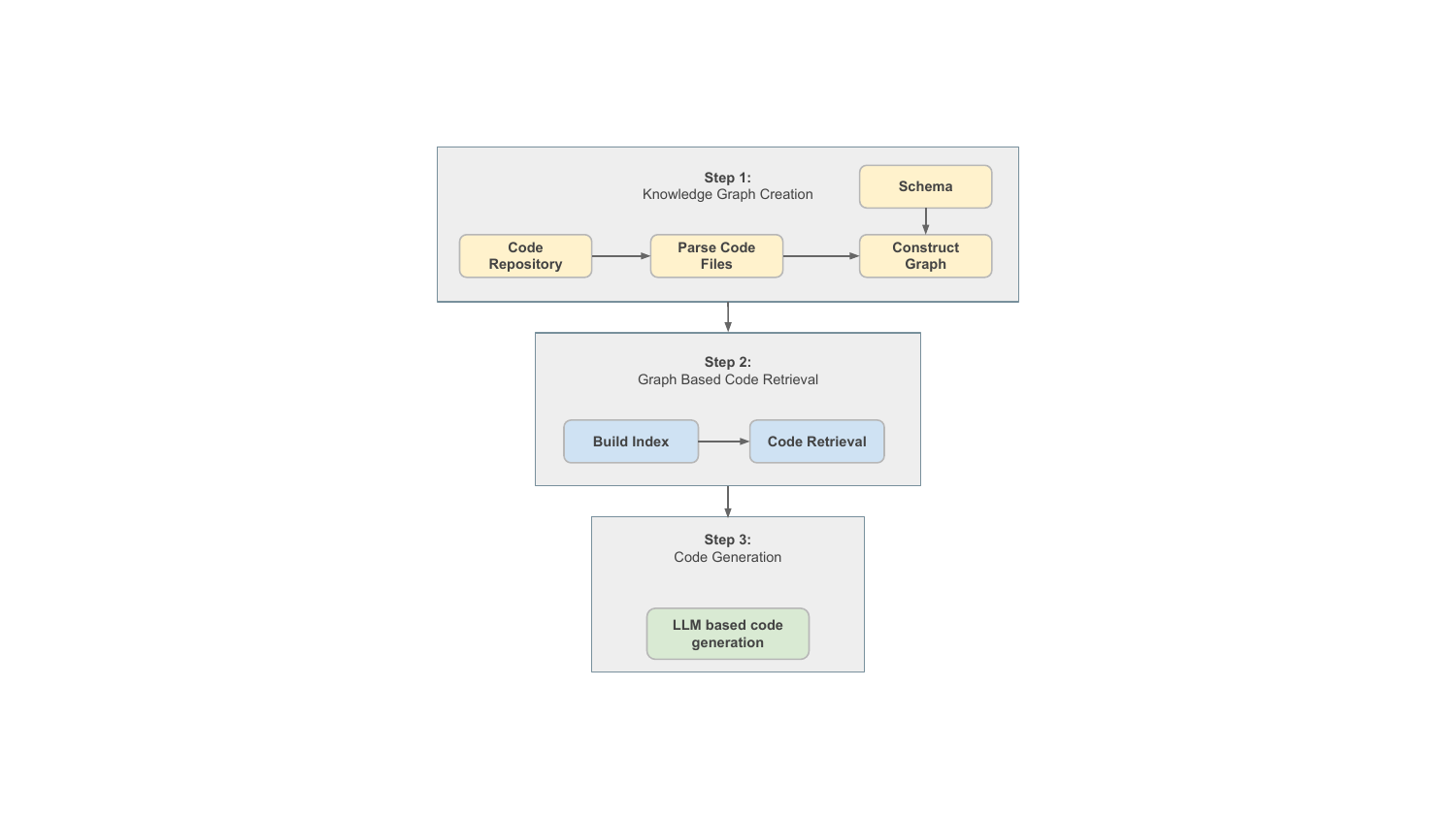}
    \caption{Three-step approach to code generation using knowledge graphs: (1) Knowledge Graph Creation: Transform the code repository into a knowledge graph, capturing connections among elements like classes, functions, etc. (2) Graph-based Code Retrieval: Create an index(knowledge base) for retrieving relevant sub-graph based on user queries, utilizes a hybrid search system (3) Code Generation with LLM: Employ a Large Language Model (LLM) to generate code, based on the retrieved sub-graph provided as context to the LLM.}
    \label{fig:arch_dig_highlevel}
\end{figure}

\par To address the aforementioned technical challenges in repository-level code generation, we propose a novel knowledge graph-based framework designed to enhance and improve code retrieval thereby improving the quality of code generation with LLMs. Our approach introduces a retrieval system capable of capturing complex dependencies and usage patterns in real-world code repositories while considering the semantic context necessary for effective downstream tasks such as code generation. We focus on adopting a hybrid approach to information retrieval on code bases. Our approach leverages a knowledge graph to represent code structures—capturing the hierarchical organization, dependencies, and usage relationships among code components like classes, functions, and modules. This structure enables context-rich interlinked searches and retrievals that support more accurate and contextually aware code generation. This relational and usage-based contextual representation of code as context to the LLM helps generate less error-prone and more functionally correct code while leveraging existing code to minimize duplication.
\par Our framework offers several technical advantages over existing approaches. By indexing code components and their relationships, the knowledge graph ensures consistent alignment of generated code with the project’s structure and requirements. Furthermore, semantic search on code documentation — enhanced by functionally relevant LLM-generated descriptions — improves retrieval accuracy by accommodating more complex user queries. This capability allows for retrieving highly relevant context that considers functional and structural similarities, resulting in efficient reuse of code snippets and improved integration with existing codebases.

\par By capturing the usage of code components such as functions and classes while providing relational information alongside code snippets to the LLM, we implicitly reflect the coding style prevalent within a specific repository context. As a result, we can generate consistently styled code for every user query, significantly reducing the manual effort required to align generated outputs.

\par In subsequent sections, we present a comprehensive analysis of our methodology, which includes knowledge graph construction for code representation, a hybrid retrieval system that combines full-text semantic searches with graph-based queries, and integration of these systems with LLMs for contextually enhanced code generation. We demonstrate our framework’s effectiveness by evaluating our framework on the EvoCodeBench\cite{li2024evocodebenchevolvingcodegeneration} dataset. We conclude by discussing the broader implications of our approach for AI-enhanced software development and outlining directions for future research in this area.

\section{Background and Related Work}

This section reviews key developments in LLM-based code generation, contextual code generation techniques, the application of knowledge graphs within software engineering, and iterative refinement methods. Furthermore, we discuss the limitations of current methodologies, highlighting how our proposed knowledge graph-based approach addresses these gaps by enhancing contextual understanding, dependency management, and consistency in style.

\par Large language models have revolutionized natural language processing tasks \cite{wei2022finetuned}\cite{chung2022scalinginstructionfinetunedlanguagemodels}\cite{openai2024gpt4technicalreport}, extending their capabilities to generating code from human language descriptions\cite{jiang2024surveylargelanguagemodels}. Seminal works, such as the introduction of Codex by Chen et al. \cite{chen2021evaluatinglargelanguagemodels}, a model fine-tuned on a large corpus of GitHub code, showcase the potential of LLMs in translating natural language prompts into functional code. Codex laid the foundation for tools like GitHub Copilot \cite{chen2021evaluatinglargelanguagemodels}, which assists developers by providing code suggestions based on natural language inputs. These models have been evaluated through benchmarks such as HumanEval \cite{chen2021evaluatinglargelanguagemodels}, which measures the ability of LLMs to generate functional code for various programming tasks, demonstrating substantial proficiency across programming languages and paradigms. However, these evaluations primarily focus on isolated code generation tasks rather than integration within existing complex codebases, limiting their practical application in real-world software development.

\par Recent research has emphasized integrating contextual information into the code generation process to address the limitations of context-free code generation. There has been a particular focus on repository-level code generation tasks\cite{pan2024enhancingrepositorylevelcodegeneration}that resemble real-world software development practices. For instance, RepoCoder\cite{zhang2023repocoderrepositorylevelcodecompletion} proposed a novel method for code generation by utilising a similarity-based retriever and a pre-trained language model for code completion. This framework aimed to extend the code context across different files in the repository by leveraging the similarity between the code to be completed and the existing code for retrieval. However, the retrieval system heavily relies on the amount of similar code in the repository for better retrieval. RepoFusion\cite{shrivastava2023repofusiontrainingcodemodels} proposes a framework that improves predictions by retrieving multiple contexts within a repository, rather than relying solely on the immediate surrounding code. Using the Fusion-in-Decoder (FiD)\cite{izacard-grave-2021-leveraging} architecture, it combines relevant repository contexts obtained through methods like prompt proposals and BM25\cite{SprckJones2000APM} scoring. While this improves the accuracy, RepoFusion faces challenges in computational scalability, limiting its deployment in resource-intensive settings.

\par In real-world settings with evolving codebases, retrieval has become a very critical aspect of improving LLM performance on code generation tasks. Research conducted by Wang et. al.\cite{wang2024coderagbenchretrievalaugmentcode} evaluates multiple retrieval and generation models on a curated dataset and shows that retrieval can significantly boost code generation performance. However, they also note that this approach works well for basic programming problems and weaker models. Furthermore, they highlight that retrieval models often struggle to find relevant documentation based on code, diminishing the actual power of retrieval.

\par Recent research has introduced agentic workflows\cite{xi2023risepotentiallargelanguage}\cite{guo2024largelanguagemodelbased} that leverage iterative refinement cycles to address some inherent limitations to single-pass code generation. AgentCoder\cite{huang2024agentcodermultiagentbasedcodegeneration}, a multi-agent framework, proposes using multiple agents with different roles to enhance code generation through effective test generation and optimization. This multi-agent system aims to reduce bias by separating code generation from testing, improving code quality and efficiency. Although powerful in generating high-quality code, multiple agents increase complexity due to coordination challenges and add overhead due to iterative refinement.

\par Knowledge graphs have emerged as powerful tools for representing complex relational information across various domains, including software engineering. GraphCoder\cite{Liu2024GraphCoderER} proposes using graphs to incorporate repository-specific knowledge to aid LLMs in code completion. It employs Code Context Graphs(CCG), which model the flow and structural relationships between code statements. During inference, the system retrieves parts of CCG to provide LLMs with better context related to the repository. However, CCGs may not adequately represent modular components of the codebase and can miss use cases requiring a semantic understanding of the knowledge base. Similarly aligned with our proposed approach, CodeXGraph\cite{liu2024codexgraphbridginglargelanguage} utilizes a knowledge graph to bridge the gap between LLMs and code repositories. It converts the repository to a knowledge graph based on a defined schema. The retrieval relies on an LLM to convert natural language questions into graph queries for retrieving relevant nodes as context for the LLM. While this method performs well for simple user queries that are easily convertible into graph queries, it struggles with complex queries requiring functional and semantic information for retrieval.

\par Despite significant strides made in code generation, retrieval, and structured knowledge representation, there remains a critical gap in integrating these approaches to address the challenges of real-world software development. Current methods lack a robust mechanism for incorporating repository-level context, managing complex dependencies, and maintaining project-specific coding standards. Our proposed knowledge graph-based framework aims to bridge this gap by representing the information within code repositories in a structured and systematic knowledge graph form. We propose using a hybrid retrieval mechanism that combines semantic, lexical, and graph-based searches to retrieve highly relevant code context in the form of sub-graphs. This enables more contextually accurate and dependency-aware code generation. By capturing key aspects such as architectural patterns, component dependencies, versioning constraints, API usage, and coding conventions, our approach provides a semantic framework that aligns generated code with existing project requirements. This reduces the need for post-generation adjustments while ensuring consistency in code style and functionality.

\par This comprehensive knowledge graph-based method contributes a novel solution to the landscape of automated code generation, addressing practical integration challenges and facilitating a more robust and context-aware development workflow. The following sections detail our methodology and system architecture, experimental results, and comparative analysis - positioning our approach as a significant advancement in the field of code retrieval and generation.

\section{Methodology} \label{methodology}

\begin{figure*}[ht]
    \centering
    \includegraphics[width=\linewidth,trim=0.2cm 1.5cm 0cm 0.0cm,clip]{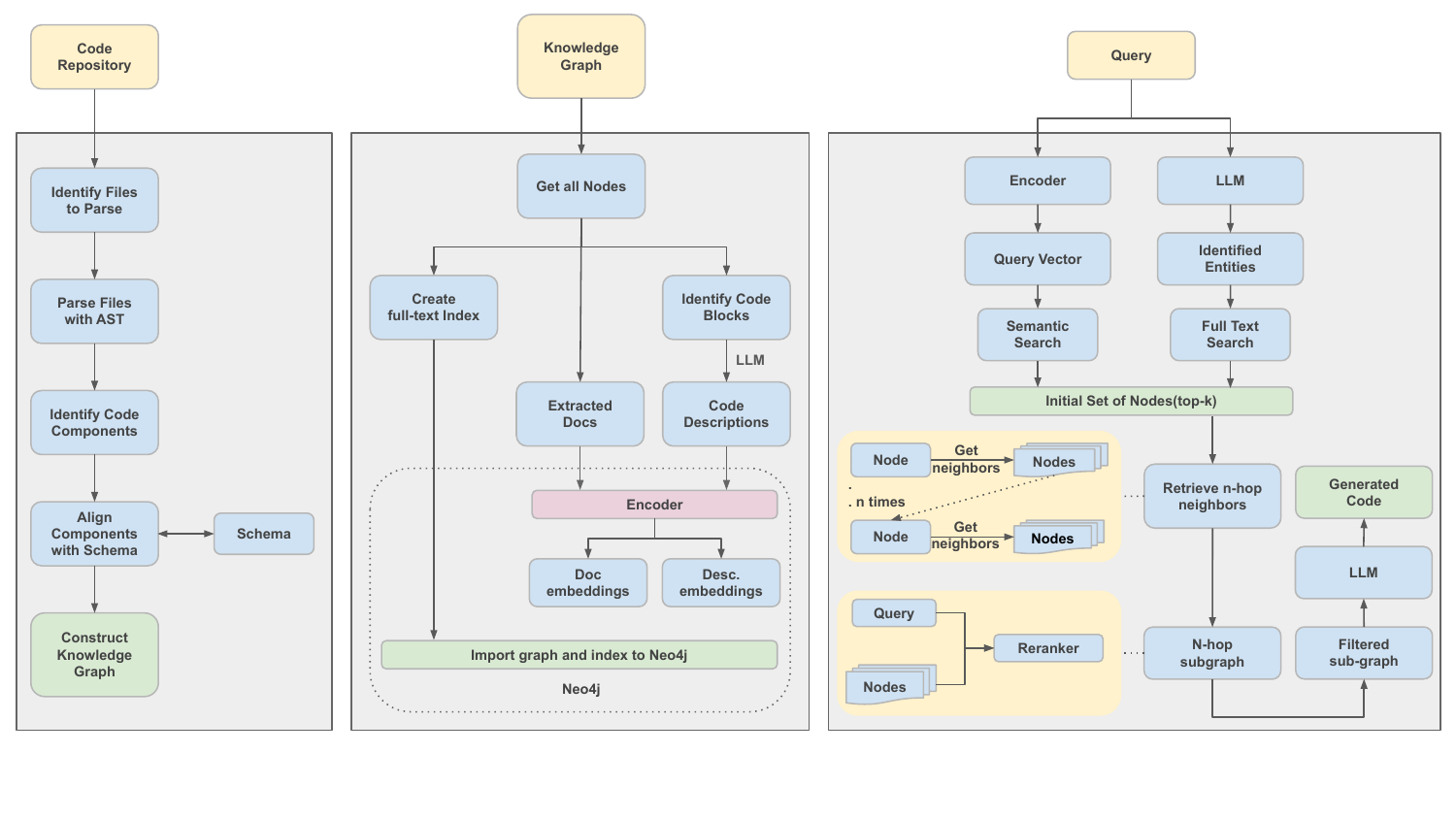}
    \caption{A diagrammatic representation of our end-to-end framework. We describe the steps in detail in the Methodology (\ref{methodology}) section.}
    \label{fig:arch_dig}
\end{figure*}

In this section, we outline our methodology for repository-level code generation. This approach is designed to enhance the contextual accuracy, integration, and relevance of generated code by grounding it within an existing codebase's structure, dependencies, and coding standards. Our methodology is organized into three main stages as shown in Fig. \ref{fig:arch_dig_highlevel}: First, constructing a comprehensive knowledge graph to represent the repository’s entities and relationships; Second, retrieving relevant code components based on a graph traversal guided by semantic relevance; and Third, using a large language model (LLM) to generate code that adheres to the repository’s architectural and stylistic constraints. Together, these stages provide a cohesive pipeline for producing high-quality, contextually appropriate code that aligns with the intricacies of the target repository. We visualize all the steps involved in our methodology in Fig. \ref{fig:arch_dig}.

\subsection{\textbf{Knowledge Graph Construction}}

In this section, we detail the process of transforming a Python code repository into an interconnected knowledge graph, which serves as the foundation for our hybrid retrieval system that retrieves relevant code snippets based on user queries.

\textbf{Parsing and Element Extraction}:
The initial step in our pipeline involves parsing each code file in the repository. To do this we use the Abstract Syntax Trees (ASTs). This systematic approach allows us to identify core elements, including \textit{Classes} ($C_i$), \textit{Methods} ($M_i$), \textit{Functions} ($F_i$), \textit{Attributes} ($A_i$), as well as other variables and dependencies.
From the elements identified by AST, we extract the essential elements that will be included in our knowledge graph. For each file, we define a set E= \{$C_i$, $M_i$, $F_i$, $A_i$, ...\} that represents the extracted elements of the code base.

\begin{figure}
    \centering
    \includegraphics[width=\linewidth,trim=4cm 0.0cm 3cm 0.0cm,clip]{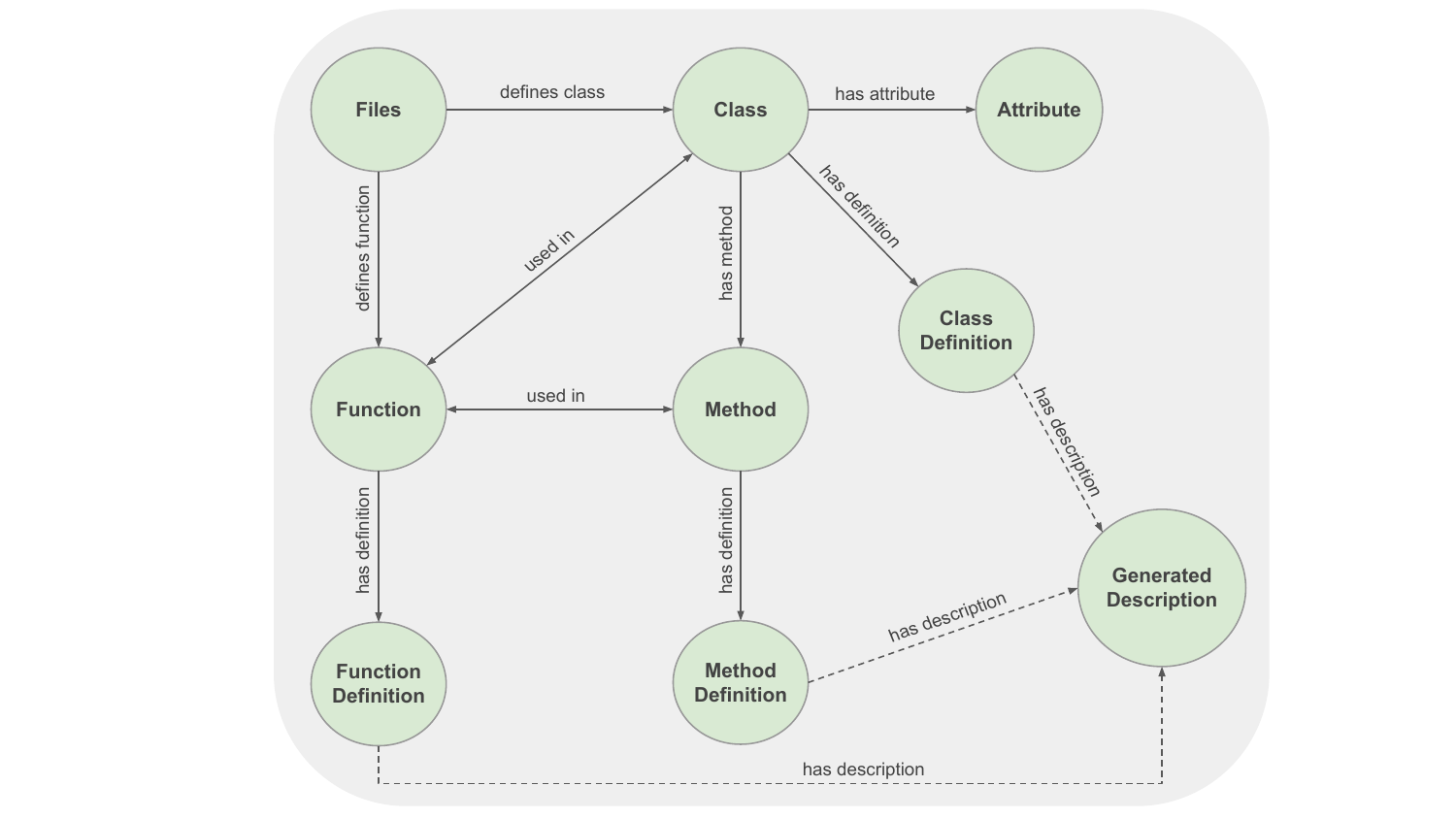}
    \caption{A sample schema that defines the node types and relations between them. Every node in the knowledge graph will be one of these types linked by relevant relation to another node.}
    \label{fig:schema}
\end{figure}

\textbf{Schema Definition}:
To provide and maintain the structure of the knowledge graph, we define the node types and relation types in a graph schema(sample shown in Fig. \ref{fig:schema}). Nodes represent distinct entities and relationships capture the interactions between these nodes. We can define the high-level schema as follows:
The set of node types is defined as V= \{File, Class, Method, Function, Attribute, Generated Description\}
and relations defined as R= \{defines class, defines a function, has a method, used in, has an attribute, has description\}. This modular schema is designed to be adaptable to various programming languages and repository types, allowing for customization based on specific requirements.

\textbf{Additional Metadata}:
In addition to the core elements, we extract documentation and comments from the code files. These are stored as additional nodes in our knowledge graph, incorporating information such as code documentation and comments. Additionally, we use an LLM to generate descriptions for code snippets to capture functional meaning and context from the code in the repository, we store these descriptions as LLM-generated descriptions in the knowledge graph. Incorporating metadata enhances the contextual depth of the knowledge graph, making it more informative and effective for retrieval tasks.

\textbf{Data Ingestion and Indexing in Neo4j}:
After successfully extracting and structuring the data according to the schema, we ingest this data into a Neo4j graph database. Once the data is in Neo4j, we create various indexes to optimize search operations. We utilize Neo4j's built-in capabilities to set up the following indexes:

\begin{itemize} 
\item \textbf{Full-text Indexes}: Function Names, Class Names, Method Names, Modules
\item \textbf{Vector Indexes}: Documentation, LLM-Generated Descriptions
\end{itemize}

We employ an encoder model (all-Mini-LM V6\cite{wang2020minilmdeepselfattentiondistillation}) to generate embeddings for the documentation and description data. These embeddings are then ingested into Neo4j's vector database to create the search index.

\textbf{Hybrid Retrieval Support}:
The full-text and vector indexes we create facilitate hybrid search capabilities, allowing for rapid retrieval based on structural elements as well as context-based retrieval through semantic search on the documentation and descriptions stored within the knowledge graph. This integrated approach enhances the overall efficiency and effectiveness of our retrieval system.

\subsection{\textbf{Hybrid Code Retrieval}}

Following knowledge graph construction, the next phase, \textit{Hybrid Code Retrieval}, focuses on identifying relevant code segments in response to user queries. We design our retrieval system to leverage the power of syntactic, semantic and graph queries. We utilize the following approach for code retrieval.

\textbf{Query Processing}: The system is designed to enable the user to enter a query in natural language. To process the query entered by the user we take two approaches. First, we use an LLM to identify specific entities if mentioned in the query. We take the query and pass it to an LLM along with the graph data schema and prompt it to identify elements from the user query that align with our schema elements. Second, we use the same encoder model used to generate embeddings for metadata to generate an embedding for the user query.

\textbf{Initial Retrieval}: Once we have the identified entities and the query embedding we perform an initial search over the knowledge graph. We perform a full-text search for the LLM-identified entities over the full-text indexes to retrieve the matching nodes from the knowledge graph. We also perform a similarity search over the vector indexes to find semantically relevant nodes to the user query. We reverse map these nodes(description and documentation) to the nodes they describe to get the functions, classes, methods etc. We limit the number of nodes to top-k for each search based on a score threshold so that we get a quality set of initially retrieved nodes.

\textbf{Expanding and Enriching Initial Set}: To enrich the initial search results, we perform a graph traversal to expand the initial nodes into a relevant sub-graph, capturing additional context around the initial matches. The depth and breadth of this traversal are governed by a hyper-parameter defining the number of hops from the starting nodes, which adjusts the extent of exploration around the query’s focal nodes. For each of the nodes in the initial subset, we do a n-hop traversal to find the dependencies and usages associated with that particular node. We call this expanded graph an n-hop sub-graph.

\textbf{Filtering Sub-graph}:
The resulting sub-graph encompasses not only the initially matched nodes but also interconnected entities, such as related functions, classes, and imports, forming a contextual framework around the core query elements. We refine this sub-graph further through semantic ranking, prioritizing nodes that align most closely with the query’s purpose while filtering out less relevant nodes. We generate embeddings for each of the nodes in the retrieved sub-graph and based on the similarity with the original query's embedding, we filter the sub-graph by keeping top-k nodes. This pruning process is essential to streamline the sub-graph, ensuring that only contextually pertinent information is passed to the LLM. This improves both computational efficiency and the model’s ability to generate contextually relevant code.

\subsection{\textbf{Code Generation with LLM}}

In the final stage of code generation, the refined sub-graph is provided as context to the LLM for code generation along with the user's initial query. The sub-graph encapsulates not only individual code snippets but also the structural and relational context surrounding these snippets in the form of the relationships between the nodes in the knowledge graph. This provides the LLM with a detailed view of the repository’s architecture, dependencies, and coding conventions, all in relevance to the user's query.
We write a custom prompt to guide the LLM, which includes instructions to utilize the relationships, dependencies, and contextual information embedded within the sub-graph. The prompt directs the LLM to focus on producing code aligned with the user's query and grounded with relevance to the existing code base. 
By grounding the LLM’s code generation in a semantically enriched sub-graph, this methodology enables the model to generate code that not only fulfils the functional requirements of the query but is also fully aligned with the repository’s framework. This knowledge graph-driven approach allows the LLM to produce contextually aware, structurally consistent code, aligned with the coding conventions and dependencies of the base repository. This holistic integration significantly reduces the need for post-generation adjustments, providing a streamlined pipeline for generating high-quality repository-specific code. Together, these stages establish a robust system for context-aware code generation, making the process more precise, efficient, and applicable to real-world development scenarios.

\begin{table}[hbt]
\renewcommand{\arraystretch}{1.3}
\caption{Comparative Analysis of pass@1 Scores Achieved by Various Code Generation Methodologies on the EvoCodeBench Dataset}
\resizebox{\columnwidth}{!}{ 

    \begin{tabular}{|c|c|c|c|}
    \hline
    \textbf{Category} & \textbf{Method} & \textbf{LLM Model} & \textbf{pass@1 Score} \\
    \hline
    \multirow{3}{*}{\begin{tabular}[c]{@{}c@{}}\textbf{Proposed}\\\textbf{Approach}\end{tabular}} 
    & \multirow{3}{*}{\begin{tabular}[c]{@{}c@{}}Graph-based\\Retrieval\end{tabular}} 
    & GPT-4 & 32.00\% \\
    \cline{3-4}
    & & GPT-4o & 33.45\% \\
    \cline{3-4}
    & & \begin{tabular}[c]{@{}c@{}}Claude 3.5\\Sonnet\end{tabular} & \textbf{36.36\%} \\
    \hline
    \multirow{6}{*}{\begin{tabular}[c]{@{}c@{}}\textbf{Baseline}\\\textbf{(EvoCodeBench)}\end{tabular}} 
    & \multirow{2}{*}{\begin{tabular}[c]{@{}c@{}}Local File\\(Infilling)\end{tabular}} 
    & GPT-4 & 20.73\% \\
    \cline{3-4}
    & & GPT-3.5 & 17.82\% \\
    \cline{2-4}
    & \multirow{2}{*}{\begin{tabular}[c]{@{}c@{}}Local File\\(Completion)\end{tabular}} 
    & GPT-4 & 17.45\% \\
    \cline{3-4}
    & & GPT-3.5 & 15.64\% \\
    \cline{2-4}
    & \multirow{2}{*}{\begin{tabular}[c]{@{}c@{}}Without\\Context\end{tabular}} 
    & GPT-4 & 7.27\% \\
    \cline{3-4}
    & & GPT-3.5 & 6.55\% \\
    \hline
    \multirow{3}{*}{\textbf{CodeXGraph}} 
    & \multirow{3}{*}{\begin{tabular}[c]{@{}c@{}}Graph-based\\(212/275 Samples)\end{tabular}} 
    & Qwen2 & 19.34\% \\
    \cline{3-4}
    & & DS-Coder & 25.47\% \\
    \cline{3-4}
    & & GPT-4o & 36.02\% \\
    \hline
    \end{tabular}
}
\label{tab:summary}
\end{table}

\section{Dataset and Evaluation}

\subsection{\textbf{Dataset: EvoCodeBench}}

To evaluate our knowledge graph-based code generation methodology, we utilized the EvoCodeBench dataset, a benchmark specifically designed to assess repository-level code generation capabilities. EvoCodeBench consists of 275 samples derived from 25 open-source repositories, selected to provide a diverse set of real-world code generation tasks. We only evaluate our approach for code written in Python programming language. Each sample corresponds to a function or method, intending to generate its code body, based on contextual information from the repository. This includes details like the namespace, location within the repository, expected functionality, and required arguments. The dataset is designed to assess models on realistic coding tasks, such as handling dependencies, integrating with existing codebases, and meeting specified functional requirements.

\subsection{\textbf{Evaluation Methodology}}

To evaluate the EvoCodeBench dataset, we slightly modify our pipeline to focus on generating the body of a function or method given the specified namespace, location, and functionality. 

Based on the given information, for each sample, we locate the function or method in the repository and replace the original body of the function with just a simple pass keyword. Following the methodology outlined in Section 3, we construct a knowledge graph of the repository, capturing essential elements and relationships. Once we have created the graph we retrieve a two-hop sub-graph starting from the target node that we want to generate the body for. We eliminate some steps from the original pipeline as we are not dealing with an actual natural language query. We already know our target and hence we only perform the graph-based retrieval for evaluation. This expansion captures the function’s immediate context and dependencies that might impact its functionality or integration with other components. The resulting sub-graph, containing nodes and edges represents these dependencies and relationships, which is then passed to the LLM as contextual input for code generation. The generated function or method body is subsequently inserted into the original file, as the body of the target function.

To evaluate the accuracy and functionality of the generated code, we utilize the pass@k metric, a standard metric in code generation benchmarks where $k$ represents the number of responses sampled from the LLM. For each sample, the EvoCodeBench dataset includes a set of predefined test cases designed to validate the generated code’s functionality. We execute these test cases on the generated code and calculate the pass rate, quantitatively measuring how well the model meets the functional requirements. The system's final performance is reported as the percentage of test cases passed across all samples, reflecting the model's effectiveness in generating functionally correct and contextually appropriate code. We discuss the results of this experiment in the following results section.

\section{Results}

We evaluated our knowledge graph based repository-level code generation approach on the EvoCodeBench dataset, using the pass@1 metric to measure code correctness on the first attempt. Table \ref{tab:summary} presents a comparison of our approach with established baselines from the original EvoCodeBench paper and the CodeXGraph model. The code correctness is measured by executing the provided test cases for each sample in the EvoCodeBench dataset.

Our approach achieved the highest pass@1 score of 36.36\% with Claude 3.5 Sonnet, modestly exceeding CodeXGraph’s top score of 36.02\% on GPT-4o. It is worth noting that CodeXGraph was evaluated on 212 out of 275 samples due to environment setup problems with some of the repositories as mentioned by the authors in the paper. We evaluate the approach on all 275 samples, we try to fix the issues faced while setting up the environment for the repositories to make the evaluation possible on all the instances available. This comprehensive evaluation allowed for a thorough assessment of how well our knowledge graph based retrieval integrates structural and contextual insights into code generation. OpenAI’s GPT models, tested with our approach, yielded pass@1 scores of 32.00\% for GPT-4 and 33.45\% for GPT-4o, which provided a significant improvement over the original EvoCodeBench baselines. These baseline methods, which included Local File Infilling, Completion and without Context, achieved pass@1 scores ranging from 7.27\% to 20.73\% for GPT-4, underscoring the challenges of context-agnostic code generation.

\section{Conclusion and Future Work}

In this work, we presented a knowledge graph based methodology for repository-level code generation, designed to enhance the contextual accuracy, dependency alignment, and stylistic consistency of generated code. Our approach leverages the structured relationships within a code repository to create a knowledge graph that captures inter-dependencies, code hierarchy, and project-specific patterns, which are then used to ground and contextualize code generation by large language models (LLMs). Evaluations on the EvoCodeBench dataset demonstrate that our method outperforms state-of-the-art baselines, achieving a pass@1 score of 36.36\% with Claude-3.5 Sonnet. This improvement underscores the value of incorporating repository context and structured semantic relationships into the code generation process, addressing key limitations associated with traditional LLM-driven code generation such as syntactic errors, incomplete code and missing dependencies. We showcase that leveraging a knowledge graph which provides the context of usage and dependencies related to the user's query and target functions helps the LLM to generate more correct and complete code.

Our method also demonstrates that leveraging a hybrid retrieval system helps retrieve highly relevant code snippets from the knowledge base compared to other retrieval methods discussed in the paper. The code context retrieved by this hybrid system when provided as a sub-graph to the LLM boosts the code generation ability of the LLM and helps it ground better on the retrieved data.

Despite its success, our approach also encountered challenges and highlights several avenues for future research. The process of sub-graph retrieval and filtering is computationally intensive, especially for large repositories, and finding the optimal balance between context richness and retrieval latency remains an ongoing research. Future work could focus on enhancing the schema design to include a broader array of code elements, such as decorators, variable types, and auxiliary files, which could increase the relevance of retrieved context for more complex queries. Extending the system to support multiple programming languages would further improve its adaptability and make it applicable to a wider range of software projects.

Additionally, integrating agent-based iterative workflows represents a promising direction for optimizing code generation accuracy and efficiency. By automating query formulation and dynamically adjusting retrieval depth based on query complexity, LLM agents could refine and execute code generation autonomously with minimal human intervention, reducing both time and computational cost. Task-specific fine-tuning of LLMs to interpret and generate code grounded in graph structures may also improve the model's ability to leverage repository context effectively.

Finally, the knowledge graph structure introduced in this paper could be extended to other downstream tasks in software engineering, including code debugging, documentation generation, and code translation. By adapting the knowledge graph to support these applications, future work could explore how this approach may serve as a generalized framework for context-aware AI-driven programming tools. Overall, our findings indicate that knowledge graph based code generation offers a robust solution to repository-specific code generation challenges, paving the way for further research and development in AI-enabled software development space.

\bibliographystyle{IEEEtran}
\bibliography{mybib}

\begin{thebibliography}{10}
\providecommand{\url}[1]{#1}
\csname url@samestyle\endcsname
\providecommand{\newblock}{\relax}
\providecommand{\bibinfo}[2]{#2}
\providecommand{\BIBentrySTDinterwordspacing}{\spaceskip=0pt\relax}
\providecommand{\BIBentryALTinterwordstretchfactor}{4}
\providecommand{\BIBentryALTinterwordspacing}{\spaceskip=\fontdimen2\font plus
\BIBentryALTinterwordstretchfactor\fontdimen3\font minus \fontdimen4\font\relax}
\providecommand{\BIBforeignlanguage}[2]{{%
\expandafter\ifx\csname l@#1\endcsname\relax
\typeout{** WARNING: IEEEtran.bst: No hyphenation pattern has been}%
\typeout{** loaded for the language `#1'. Using the pattern for}%
\typeout{** the default language instead.}%
\else
\language=\csname l@#1\endcsname
\fi
#2}}
\providecommand{\BIBdecl}{\relax}
\BIBdecl

\bibitem{openai2024gpt4technicalreport}
\BIBentryALTinterwordspacing
OpenAI, J.~Achiam, S.~Adler, S.~Agarwal, L.~Ahmad, I.~Akkaya, F.~L. Aleman, D.~Almeida, J.~Altenschmidt, S.~Altman, S.~Anadkat, R.~Avila, I.~Babuschkin, S.~Balaji, V.~Balcom, P.~Baltescu, H.~Bao, M.~Bavarian, J.~Belgum, I.~Bello, J.~Berdine, G.~Bernadett-Shapiro, C.~Berner, L.~Bogdonoff, O.~Boiko, M.~Boyd, A.-L. Brakman, G.~Brockman, T.~Brooks, M.~Brundage, K.~Button, T.~Cai, R.~Campbell, A.~Cann, B.~Carey, C.~Carlson, R.~Carmichael, B.~Chan, C.~Chang, F.~Chantzis, D.~Chen, S.~Chen, R.~Chen, J.~Chen, M.~Chen, B.~Chess, C.~Cho, C.~Chu, H.~W. Chung, D.~Cummings, J.~Currier, Y.~Dai, C.~Decareaux, T.~Degry, N.~Deutsch, D.~Deville, A.~Dhar, D.~Dohan, S.~Dowling, S.~Dunning, A.~Ecoffet, A.~Eleti, T.~Eloundou, D.~Farhi, L.~Fedus, N.~Felix, S.~P. Fishman, J.~Forte, I.~Fulford, L.~Gao, E.~Georges, C.~Gibson, V.~Goel, T.~Gogineni, G.~Goh, R.~Gontijo-Lopes, J.~Gordon, M.~Grafstein, S.~Gray, R.~Greene, J.~Gross, S.~S. Gu, Y.~Guo, C.~Hallacy, J.~Han, J.~Harris, Y.~He, M.~Heaton, J.~Heidecke, C.~Hesse, A.~Hickey,
  W.~Hickey, P.~Hoeschele, B.~Houghton, K.~Hsu, S.~Hu, X.~Hu, J.~Huizinga, S.~Jain, S.~Jain, J.~Jang, A.~Jiang, R.~Jiang, H.~Jin, D.~Jin, S.~Jomoto, B.~Jonn, H.~Jun, T.~Kaftan, Łukasz Kaiser, A.~Kamali, I.~Kanitscheider, N.~S. Keskar, T.~Khan, L.~Kilpatrick, J.~W. Kim, C.~Kim, Y.~Kim, J.~H. Kirchner, J.~Kiros, M.~Knight, D.~Kokotajlo, Łukasz Kondraciuk, A.~Kondrich, A.~Konstantinidis, K.~Kosic, G.~Krueger, V.~Kuo, M.~Lampe, I.~Lan, T.~Lee, J.~Leike, J.~Leung, D.~Levy, C.~M. Li, R.~Lim, M.~Lin, S.~Lin, M.~Litwin, T.~Lopez, R.~Lowe, P.~Lue, A.~Makanju, K.~Malfacini, S.~Manning, T.~Markov, Y.~Markovski, B.~Martin, K.~Mayer, A.~Mayne, B.~McGrew, S.~M. McKinney, C.~McLeavey, P.~McMillan, J.~McNeil, D.~Medina, A.~Mehta, J.~Menick, L.~Metz, A.~Mishchenko, P.~Mishkin, V.~Monaco, E.~Morikawa, D.~Mossing, T.~Mu, M.~Murati, O.~Murk, D.~Mély, A.~Nair, R.~Nakano, R.~Nayak, A.~Neelakantan, R.~Ngo, H.~Noh, L.~Ouyang, C.~O'Keefe, J.~Pachocki, A.~Paino, J.~Palermo, A.~Pantuliano, G.~Parascandolo, J.~Parish, E.~Parparita,
  A.~Passos, M.~Pavlov, A.~Peng, A.~Perelman, F.~de~Avila Belbute~Peres, M.~Petrov, H.~P. de~Oliveira~Pinto, Michael, Pokorny, M.~Pokrass, V.~H. Pong, T.~Powell, A.~Power, B.~Power, E.~Proehl, R.~Puri, A.~Radford, J.~Rae, A.~Ramesh, C.~Raymond, F.~Real, K.~Rimbach, C.~Ross, B.~Rotsted, H.~Roussez, N.~Ryder, M.~Saltarelli, T.~Sanders, S.~Santurkar, G.~Sastry, H.~Schmidt, D.~Schnurr, J.~Schulman, D.~Selsam, K.~Sheppard, T.~Sherbakov, J.~Shieh, S.~Shoker, P.~Shyam, S.~Sidor, E.~Sigler, M.~Simens, J.~Sitkin, K.~Slama, I.~Sohl, B.~Sokolowsky, Y.~Song, N.~Staudacher, F.~P. Such, N.~Summers, I.~Sutskever, J.~Tang, N.~Tezak, M.~B. Thompson, P.~Tillet, A.~Tootoonchian, E.~Tseng, P.~Tuggle, N.~Turley, J.~Tworek, J.~F.~C. Uribe, A.~Vallone, A.~Vijayvergiya, C.~Voss, C.~Wainwright, J.~J. Wang, A.~Wang, B.~Wang, J.~Ward, J.~Wei, C.~Weinmann, A.~Welihinda, P.~Welinder, J.~Weng, L.~Weng, M.~Wiethoff, D.~Willner, C.~Winter, S.~Wolrich, H.~Wong, L.~Workman, S.~Wu, J.~Wu, M.~Wu, K.~Xiao, T.~Xu, S.~Yoo, K.~Yu, Q.~Yuan,
  W.~Zaremba, R.~Zellers, C.~Zhang, M.~Zhang, S.~Zhao, T.~Zheng, J.~Zhuang, W.~Zhuk, and B.~Zoph, ``Gpt-4 technical report,'' 2024. [Online]. Available: \url{https://arxiv.org/abs/2303.08774}
\BIBentrySTDinterwordspacing

\bibitem{chen2021evaluatinglargelanguagemodels}
\BIBentryALTinterwordspacing
M.~Chen, J.~Tworek, H.~Jun, Q.~Yuan, H.~P. de~Oliveira~Pinto, J.~Kaplan, H.~Edwards, Y.~Burda, N.~Joseph, G.~Brockman, A.~Ray, R.~Puri, G.~Krueger, M.~Petrov, H.~Khlaaf, G.~Sastry, P.~Mishkin, B.~Chan, S.~Gray, N.~Ryder, M.~Pavlov, A.~Power, L.~Kaiser, M.~Bavarian, C.~Winter, P.~Tillet, F.~P. Such, D.~Cummings, M.~Plappert, F.~Chantzis, E.~Barnes, A.~Herbert-Voss, W.~H. Guss, A.~Nichol, A.~Paino, N.~Tezak, J.~Tang, I.~Babuschkin, S.~Balaji, S.~Jain, W.~Saunders, C.~Hesse, A.~N. Carr, J.~Leike, J.~Achiam, V.~Misra, E.~Morikawa, A.~Radford, M.~Knight, M.~Brundage, M.~Murati, K.~Mayer, P.~Welinder, B.~McGrew, D.~Amodei, S.~McCandlish, I.~Sutskever, and W.~Zaremba, ``Evaluating large language models trained on code,'' 2021. [Online]. Available: \url{https://arxiv.org/abs/2107.03374}
\BIBentrySTDinterwordspacing

\bibitem{10109345}
I.~Ozkaya, ``Application of large language models to software engineering tasks: Opportunities, risks, and implications,'' \emph{IEEE Software}, vol.~40, no.~3, pp. 4--8, 2023.

\bibitem{dou2024whatswrongcodegenerated}
\BIBentryALTinterwordspacing
S.~Dou, H.~Jia, S.~Wu, H.~Zheng, W.~Zhou, M.~Wu, M.~Chai, J.~Fan, C.~Huang, Y.~Tao, Y.~Liu, E.~Zhou, M.~Zhang, Y.~Zhou, Y.~Wu, R.~Zheng, M.~Wen, R.~Weng, J.~Wang, X.~Cai, T.~Gui, X.~Qiu, Q.~Zhang, and X.~Huang, ``What's wrong with your code generated by large language models? an extensive study,'' 2024. [Online]. Available: \url{https://arxiv.org/abs/2407.06153}
\BIBentrySTDinterwordspacing

\bibitem{hou2024largelanguagemodelssoftware}
\BIBentryALTinterwordspacing
X.~Hou, Y.~Zhao, Y.~Liu, Z.~Yang, K.~Wang, L.~Li, X.~Luo, D.~Lo, J.~Grundy, and H.~Wang, ``Large language models for software engineering: A systematic literature review,'' 2024. [Online]. Available: \url{https://arxiv.org/abs/2308.10620}
\BIBentrySTDinterwordspacing

\bibitem{wang2024functionalcorrectnessinvestigatingcoding}
\BIBentryALTinterwordspacing
Y.~Wang, T.~Jiang, M.~Liu, J.~Chen, and Z.~Zheng, ``Beyond functional correctness: Investigating coding style inconsistencies in large language models,'' 2024. [Online]. Available: \url{https://arxiv.org/abs/2407.00456}
\BIBentrySTDinterwordspacing

\bibitem{Di_Grazia_2023}
\BIBentryALTinterwordspacing
L.~Di~Grazia and M.~Pradel, ``Code search: A survey of techniques for finding code,'' \emph{ACM Computing Surveys}, vol.~55, no.~11, p. 1–31, Feb. 2023. [Online]. Available: \url{http://dx.doi.org/10.1145/3565971}
\BIBentrySTDinterwordspacing

\bibitem{tan2024promptbasedcodecompletionmultiretrieval}
\BIBentryALTinterwordspacing
H.~Tan, Q.~Luo, L.~Jiang, Z.~Zhan, J.~Li, H.~Zhang, and Y.~Zhang, ``Prompt-based code completion via multi-retrieval augmented generation,'' 2024. [Online]. Available: \url{https://arxiv.org/abs/2405.07530}
\BIBentrySTDinterwordspacing

\bibitem{NEURIPS2020_6b493230}
\BIBentryALTinterwordspacing
P.~Lewis, E.~Perez, A.~Piktus, F.~Petroni, V.~Karpukhin, N.~Goyal, H.~K\"{u}ttler, M.~Lewis, W.-t. Yih, T.~Rockt\"{a}schel, S.~Riedel, and D.~Kiela, ``Retrieval-augmented generation for knowledge-intensive nlp tasks,'' in \emph{Advances in Neural Information Processing Systems}, H.~Larochelle, M.~Ranzato, R.~Hadsell, M.~Balcan, and H.~Lin, Eds., vol.~33.\hskip 1em plus 0.5em minus 0.4em\relax Curran Associates, Inc., 2020, pp. 9459--9474. [Online]. Available: \url{https://proceedings.neurips.cc/paper_files/paper/2020/file/6b493230205f780e1bc26945df7481e5-Paper.pdf}
\BIBentrySTDinterwordspacing

\bibitem{li2024reviewprominentparadigmsllmbased}
\BIBentryALTinterwordspacing
X.~Li, ``A review of prominent paradigms for llm-based agents: Tool use (including rag), planning, and feedback learning,'' 2024. [Online]. Available: \url{https://arxiv.org/abs/2406.05804}
\BIBentrySTDinterwordspacing

\bibitem{nguyen2024agilecoderdynamiccollaborativeagents}
\BIBentryALTinterwordspacing
M.~H. Nguyen, T.~P. Chau, P.~X. Nguyen, and N.~D.~Q. Bui, ``Agilecoder: Dynamic collaborative agents for software development based on agile methodology,'' 2024. [Online]. Available: \url{https://arxiv.org/abs/2406.11912}
\BIBentrySTDinterwordspacing

\bibitem{li2024evocodebenchevolvingcodegeneration}
\BIBentryALTinterwordspacing
J.~Li, G.~Li, X.~Zhang, Y.~Dong, and Z.~Jin, ``Evocodebench: An evolving code generation benchmark aligned with real-world code repositories,'' 2024. [Online]. Available: \url{https://arxiv.org/abs/2404.00599}
\BIBentrySTDinterwordspacing

\bibitem{wei2022finetuned}
\BIBentryALTinterwordspacing
J.~Wei, M.~Bosma, V.~Zhao, K.~Guu, A.~W. Yu, B.~Lester, N.~Du, A.~M. Dai, and Q.~V. Le, ``Finetuned language models are zero-shot learners,'' in \emph{International Conference on Learning Representations}, 2022. [Online]. Available: \url{https://openreview.net/forum?id=gEZrGCozdqR}
\BIBentrySTDinterwordspacing

\bibitem{chung2022scalinginstructionfinetunedlanguagemodels}
\BIBentryALTinterwordspacing
H.~W. Chung, L.~Hou, S.~Longpre, B.~Zoph, Y.~Tay, W.~Fedus, Y.~Li, X.~Wang, M.~Dehghani, S.~Brahma, A.~Webson, S.~S. Gu, Z.~Dai, M.~Suzgun, X.~Chen, A.~Chowdhery, A.~Castro-Ros, M.~Pellat, K.~Robinson, D.~Valter, S.~Narang, G.~Mishra, A.~Yu, V.~Zhao, Y.~Huang, A.~Dai, H.~Yu, S.~Petrov, E.~H. Chi, J.~Dean, J.~Devlin, A.~Roberts, D.~Zhou, Q.~V. Le, and J.~Wei, ``Scaling instruction-finetuned language models,'' 2022. [Online]. Available: \url{https://arxiv.org/abs/2210.11416}
\BIBentrySTDinterwordspacing

\bibitem{jiang2024surveylargelanguagemodels}
\BIBentryALTinterwordspacing
J.~Jiang, F.~Wang, J.~Shen, S.~Kim, and S.~Kim, ``A survey on large language models for code generation,'' 2024. [Online]. Available: \url{https://arxiv.org/abs/2406.00515}
\BIBentrySTDinterwordspacing

\bibitem{pan2024enhancingrepositorylevelcodegeneration}
\BIBentryALTinterwordspacing
Z.~Pan, X.~Hu, X.~Xia, and X.~Yang, ``Enhancing repository-level code generation with integrated contextual information,'' 2024. [Online]. Available: \url{https://arxiv.org/abs/2406.03283}
\BIBentrySTDinterwordspacing

\bibitem{zhang2023repocoderrepositorylevelcodecompletion}
\BIBentryALTinterwordspacing
F.~Zhang, B.~Chen, Y.~Zhang, J.~Keung, J.~Liu, D.~Zan, Y.~Mao, J.-G. Lou, and W.~Chen, ``Repocoder: Repository-level code completion through iterative retrieval and generation,'' 2023. [Online]. Available: \url{https://arxiv.org/abs/2303.12570}
\BIBentrySTDinterwordspacing

\bibitem{shrivastava2023repofusiontrainingcodemodels}
\BIBentryALTinterwordspacing
D.~Shrivastava, D.~Kocetkov, H.~de~Vries, D.~Bahdanau, and T.~Scholak, ``Repofusion: Training code models to understand your repository,'' 2023. [Online]. Available: \url{https://arxiv.org/abs/2306.10998}
\BIBentrySTDinterwordspacing

\bibitem{izacard-grave-2021-leveraging}
\BIBentryALTinterwordspacing
G.~Izacard and E.~Grave, ``Leveraging passage retrieval with generative models for open domain question answering,'' in \emph{Proceedings of the 16th Conference of the European Chapter of the Association for Computational Linguistics: Main Volume}, P.~Merlo, J.~Tiedemann, and R.~Tsarfaty, Eds.\hskip 1em plus 0.5em minus 0.4em\relax Online: Association for Computational Linguistics, Apr. 2021, pp. 874--880. [Online]. Available: \url{https://aclanthology.org/2021.eacl-main.74}
\BIBentrySTDinterwordspacing

\bibitem{SprckJones2000APM}
\BIBentryALTinterwordspacing
K.~S. Jones, S.~Walker, and S.~E. Robertson, ``A probabilistic model of information retrieval: development and comparative experiments - part 1,'' \emph{Inf. Process. Manag.}, vol.~36, pp. 779--808, 2000. [Online]. Available: \url{https://api.semanticscholar.org/CorpusID:1965284}
\BIBentrySTDinterwordspacing

\bibitem{wang2024coderagbenchretrievalaugmentcode}
\BIBentryALTinterwordspacing
Z.~Z. Wang, A.~Asai, X.~V. Yu, F.~F. Xu, Y.~Xie, G.~Neubig, and D.~Fried, ``Coderag-bench: Can retrieval augment code generation?'' 2024. [Online]. Available: \url{https://arxiv.org/abs/2406.14497}
\BIBentrySTDinterwordspacing

\bibitem{xi2023risepotentiallargelanguage}
\BIBentryALTinterwordspacing
Z.~Xi, W.~Chen, X.~Guo, W.~He, Y.~Ding, B.~Hong, M.~Zhang, J.~Wang, S.~Jin, E.~Zhou, R.~Zheng, X.~Fan, X.~Wang, L.~Xiong, Y.~Zhou, W.~Wang, C.~Jiang, Y.~Zou, X.~Liu, Z.~Yin, S.~Dou, R.~Weng, W.~Cheng, Q.~Zhang, W.~Qin, Y.~Zheng, X.~Qiu, X.~Huang, and T.~Gui, ``The rise and potential of large language model based agents: A survey,'' 2023. [Online]. Available: \url{https://arxiv.org/abs/2309.07864}
\BIBentrySTDinterwordspacing

\bibitem{guo2024largelanguagemodelbased}
\BIBentryALTinterwordspacing
T.~Guo, X.~Chen, Y.~Wang, R.~Chang, S.~Pei, N.~V. Chawla, O.~Wiest, and X.~Zhang, ``Large language model based multi-agents: A survey of progress and challenges,'' 2024. [Online]. Available: \url{https://arxiv.org/abs/2402.01680}
\BIBentrySTDinterwordspacing

\bibitem{huang2024agentcodermultiagentbasedcodegeneration}
\BIBentryALTinterwordspacing
D.~Huang, J.~M. Zhang, M.~Luck, Q.~Bu, Y.~Qing, and H.~Cui, ``Agentcoder: Multi-agent-based code generation with iterative testing and optimisation,'' 2024. [Online]. Available: \url{https://arxiv.org/abs/2312.13010}
\BIBentrySTDinterwordspacing

\bibitem{Liu2024GraphCoderER}
\BIBentryALTinterwordspacing
W.~Liu, A.~Yu, D.~Zan, B.~Shen, W.~Zhang, H.~Zhao, Z.~Jin, and Q.~Wang, ``Graphcoder: Enhancing repository-level code completion via code context graph-based retrieval and language model,'' \emph{ArXiv}, vol. abs/2406.07003, 2024. [Online]. Available: \url{https://api.semanticscholar.org/CorpusID:270379479}
\BIBentrySTDinterwordspacing

\bibitem{liu2024codexgraphbridginglargelanguage}
\BIBentryALTinterwordspacing
X.~Liu, B.~Lan, Z.~Hu, Y.~Liu, Z.~Zhang, F.~Wang, M.~Shieh, and W.~Zhou, ``Codexgraph: Bridging large language models and code repositories via code graph databases,'' 2024. [Online]. Available: \url{https://arxiv.org/abs/2408.03910}
\BIBentrySTDinterwordspacing

\bibitem{wang2020minilmdeepselfattentiondistillation}
\BIBentryALTinterwordspacing
W.~Wang, F.~Wei, L.~Dong, H.~Bao, N.~Yang, and M.~Zhou, ``Minilm: Deep self-attention distillation for task-agnostic compression of pre-trained transformers,'' 2020. [Online]. Available: \url{https://arxiv.org/abs/2002.10957}
\BIBentrySTDinterwordspacing

\end{thebibliography}

\end{document}